# Indonesian Automatic Speech Recognition with XLSR-53

Panji Arisaputra*, Amalia Zahra
Computer Science Department, BINUS Graduate Program, Master of Computer Science, Bina Nusantara University, Jakarta, 11480, Indonesia

Corresponding Author Email: panji.arisaputra@binus.ac.id

***Keywords:***
*Indonesian, Automatic Speech Recognition, wav2vec 2.0, XLSR-53*

**ABSTRACT**
This study focuses on the development of Indonesian Automatic Speech Recognition (ASR) using the XLSR-53 pre-trained model, the XLSR stands for cross-lingual speech representations. The use of this XLSR-53 pre-trained model is to significantly reduce the amount of training data in non-English languages required to achieve a competitive Word Error Rate (WER). The total amount of data used in this study is 24 hours, 18 minutes, and 1 second: (1) TITML-IDN 14 hours and 31 minutes; (2) Magic Data 3 hours and 33 minutes; and (3) Common Voice 6 hours, 14 minutes, and 1 second. With a WER of 20%, the model built in this study can compete with similar models using the Common Voice dataset split test. WER can be decreased by around 8% using a language model, resulted in WER from 20% to 12%. Thus, the results of this study have succeeded in perfecting previous research in contributing to the creation of a better Indonesian ASR with a smaller amount of data.

## 1. INTRODUCTION

Globalization has massively tried to introduce English into all kinds of activities of everyone, there are still many people learning foreign languages and more and more trying to deepen them. The use of information technology currently is also one of the conditions for successful learning. Modern information technology should be an effective tool that will facilitate the assimilation of knowledge, making learning interactive, communicatively oriented, interesting, visual and individual. In this environment, future generations not only communicate, but also build professional relationships, position their interests and represent themselves. In conclusion, there is now what is called the generation with its head down (Kopf Unten Generation) as the Germans call today's youth because of their excessive love for smartphones. The latest technologies, as well as the growing number of modern digital media resources that serve as carriers of educational information, make lessons more flexible, learning processes more autonomous [1].

ASR is a technology that is programmed to automatically transcribe spoken language into text (speech-to-text). Typical use of ASR has a model that minimizes the WER metric when transcribing speech input. One of the uses of ASR is that it can help better bridge the human-human (human-to-human) and human-machine (human-to-machine) communication [2].

The need for ASR around us is not a new phenomenon but has reached a more complex stage. Some examples of the application of ASR itself can be found in air traffic control, security and biometric identification, gaming, YouTube closed captioning, voicemail transcription, home automation, and many more [3].

In **Figure 1** shows that in the last five years (2016-2020) the progress of Information and Communication Technology (ICT) in Indonesia is very rapid, and the role of ICT has become a major interest in people's lives [4]. We can also see that the percentage of individuals accessing the Internet and households accessing the Internet has grown rapidly, whereas the percentage of individuals owning a mobile phone and households owning a computer has grown relatively slowly. This happens because what is meant by "accessing the internet" is the intensity of someone accessing the internet and is not related to owning a mobile phone and a computer, but is more related to software developments over the past years. Back in 2016 – 2017 as we can see in **Figure 2**, there were not so many social media and internet users, which meant that individuals rarely accessed the internet even though they had computers and mobile phones because there was not much they could do, until when social media and online games, especially mobile games, exploded. So even though ownership of computers and mobile phones did not increase significantly, the intensity of accessing the internet through both increased significantly [5, 6].

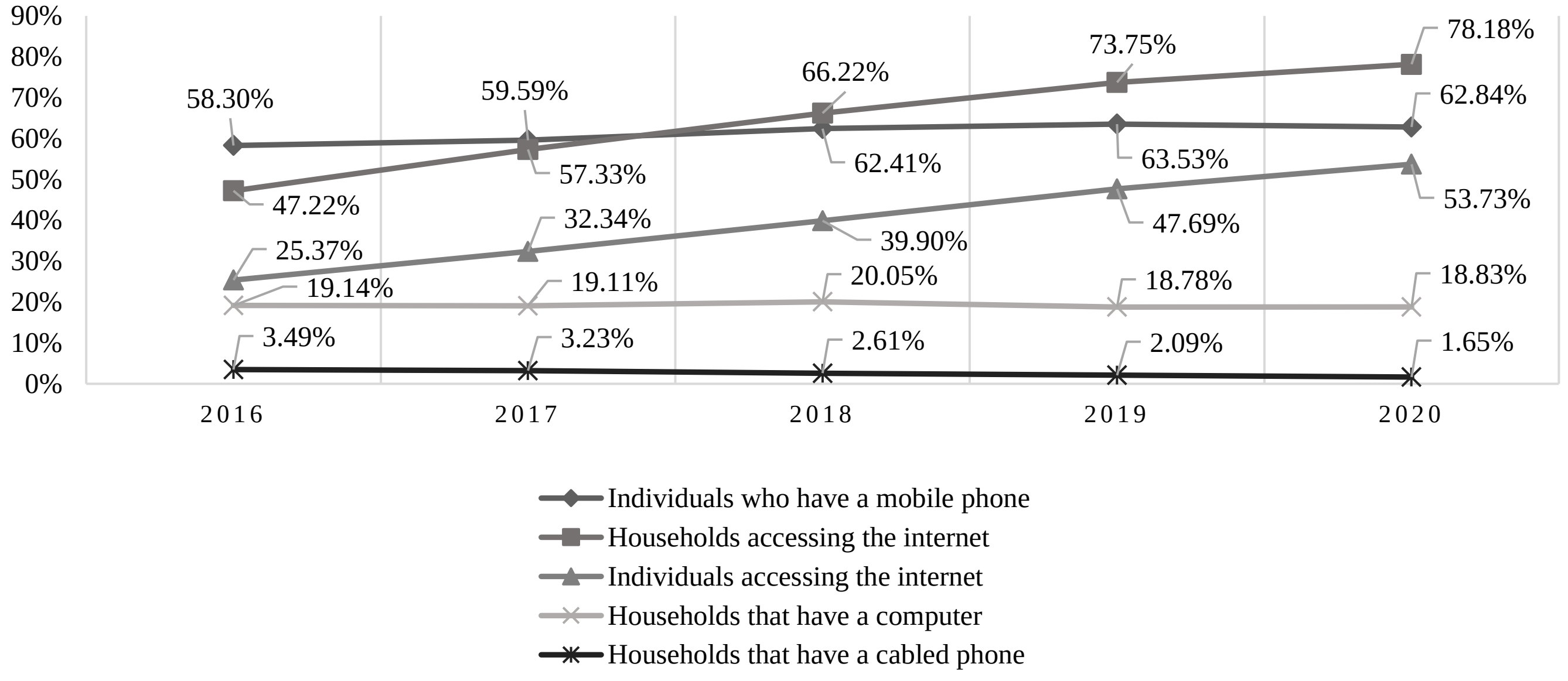

**Figure 1.** Development of ICT Indicators in Indonesia 2016-2020

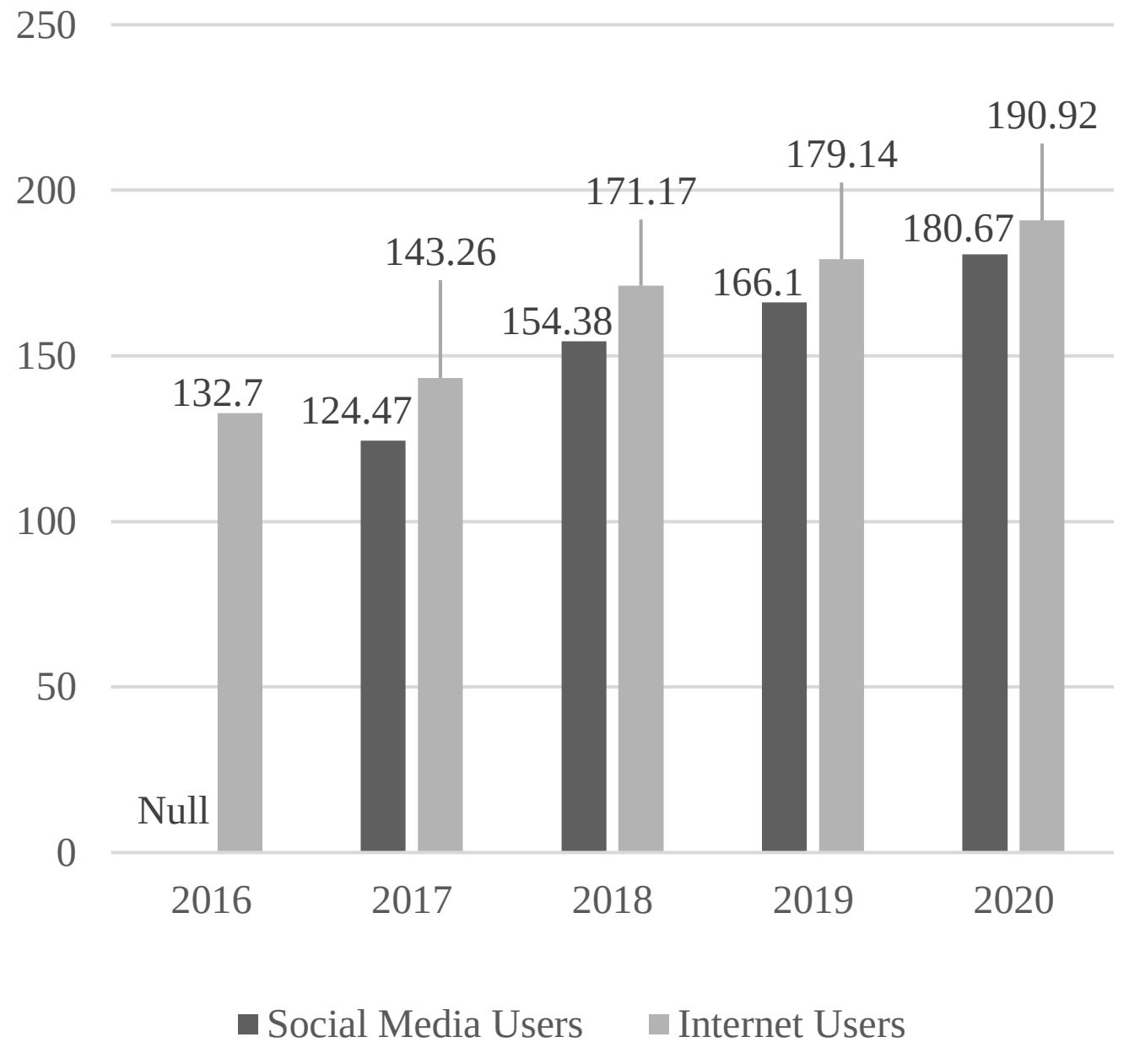

**Figure 2.** Number of Social Media and Internet Users in Indonesia 2016-2020 (in Million)

Yu & Deng [2] also revealed that speech is the most natural interaction between humans when compared to writing or typing. This is because users can experience the technology without having to see the interface of the technology directly if they have implemented the ASR system so that it can make it easier for someone who can't even read and write to use the technology. With the development of ICT advancements in Indonesia and knowing the fact that ASR has become an important modality in today's modern technology, therefore the development of ASR technology using the Indonesian language must be intensified considering the benefits offered by ASR can facilitate the activities of Indonesian people in their daily lives.

XLSR makes it possible to study speech representation across languages. The approach to XLSR is carried out by pre-training on a single model of raw speech waveforms in various languages. The experiment conducted by Conneau, et al. [7] showed that cross-language pre-training significantly outperformed monolingual pre-training. Conneau, et al. [7] also mentioned that there is an XLSR-53 which is a pre-trained model that is trained in 53 different languages on top of the wav2vec 2.0 model, which makes the XLSR-53 pre-trained model has the same architecture with the wav2vec 2.0 model. The dataset used in the model pre-training uses 56,000 hours of data sources from Mozilla's Common Voice (Common Voice), BABEL, and Multilingual LibriSpeech. All of these datasets are combined into one big dataset.

In **Figure 3**, it can be seen how latent speech representation in XLSR-53 is shared between languages. The most significant benefit when sharing this latent speech representation is the possibility to use language features from other languages without pre-training the model again in the new language. Further explanation regarding the architecture will be explained in **Section 4**.

Three datasets will be used: (1) TITML-IDN (Tokyo Institute of Technology Multilingual Speech Corpus-Indonesian); (2) Magic Data; and (3) Common Voice. This study in general is fine-tuning a pre-trained model, where it works by adjusting precisely to bring it to the highest level of performance or effectiveness, and in this study it is done by retraining the pre-trained model using new datasets (in this case using the Indonesian dataset) so that the pre-trained model can learn and understand Indonesian more effectively than any other language, considering that the XLSR-53 pre-trained model has been trained in 53 different languages. In addition, this study also tries to contribute to a new type of recipe when using a pre-trainied model, where before entering the fine-tuning step, it first combines the dataset first, then divides it into training, validation, and testing. The combination of the three datasets will be tested with different pairs from each other, such as: (1) using one dataset at a time; (2) combine two datasets with each other in every possibility; and (3) combine all three datasets. In addition, each combination will also be tested with different language model settings, such as: (1) not using a language model; (2) with 2-gram KenLM; (3) with 3-gram KenLM; (4) with 4-gram KenLM; and (5) with 5-gram KenLM.

The test split of the Common Voice dataset is used for evaluation or inference of the designed model, this dataset will later become a benchmark in performance comparisons with the WER metric. All variation results will be compared and reported.

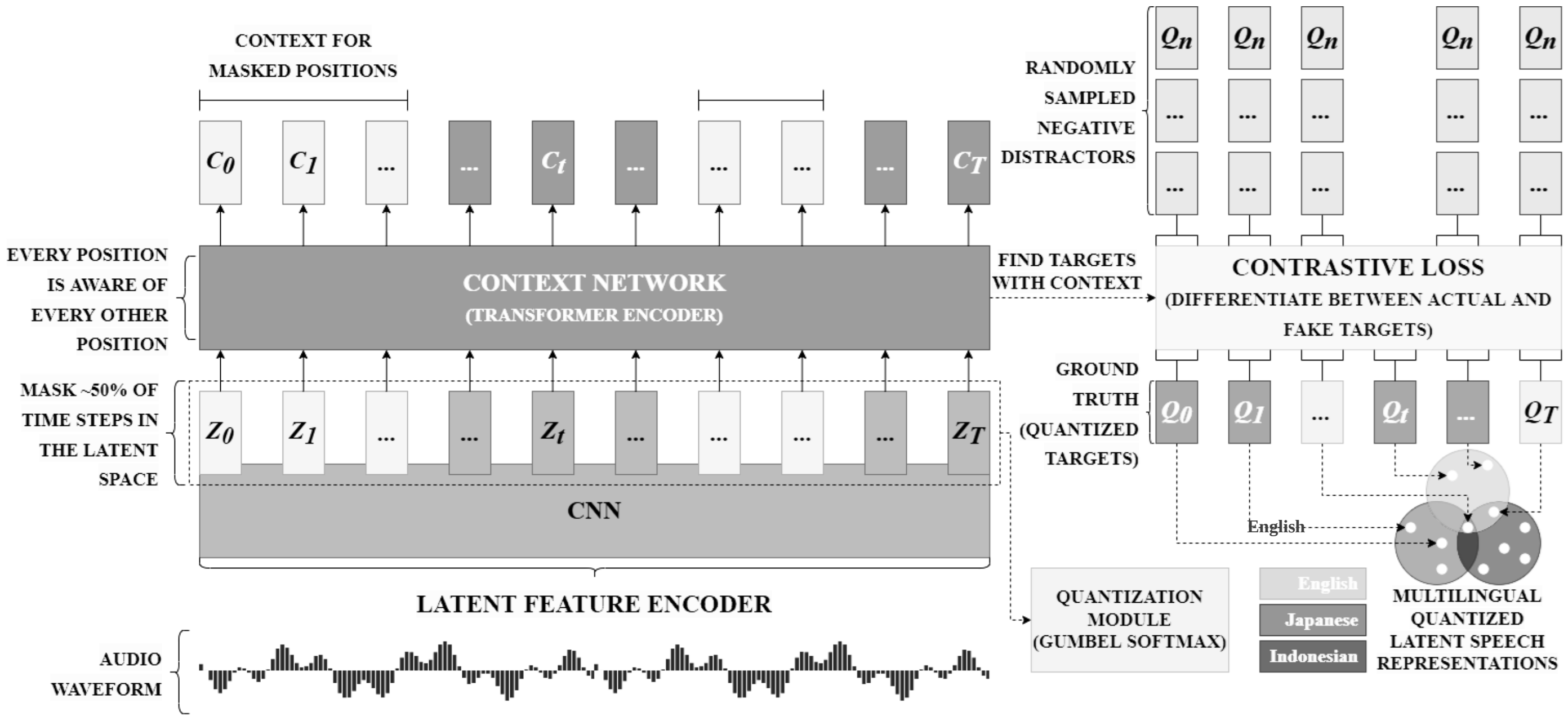

**Figure 3.** XLSR Architecture

## 2. RELATED WORK

**Wav2vec 2.0**. This model is a combination of previous models such as CPC (Contrastive Predictive Coding) [8], MPC (Model Predictive Control) [9], wav2vec [10], and vq-wav2vec [11]. This model studies the robust representation of speech in an unsupervised manner by predicting the correct speech unit for the masked audio portion. Similar to the previous model, this model combines CNN (Convolutional Neural Network) as a feature encoder, MPC as a modeling language, and contrastive tasks such as CPC. This model outperformed previous models such as wav2vec and vq-wav2vec [12].

**XLSR-53**. Another alternative to the wav2vec 2.0 model is the XLSR-53 [7]. Self-supervised learning in this alternative model can learn cross-lingual speech representations by training a single model of the raw waveform in various languages, including Indonesian. Although this model is built on the wav2vec 2.0 model which is trained by completing a contrastive task on masked latent speech representations, the XLSR-53 can learn latent quantization spread across languages. Architectural differences are also found in quantization based on product quantization by selecting quantized representations from codebooks. Gumbel-Softmax allows the selection of discrete codebook entries in a completely distinguishable manner. The context network architecture in this model is similar to BERT (Bidirectional Encoder Representations from Transformers) [13], with the difference being in the relative positional embedding steps, the goal of which requires identifying the true quantized latent for the masked time-step in a set of $K$ = 100 distractors sampled from other masked time-steps. The trade-offs are similar in text comprehension: multilingual pretraining can significantly improve performance in low-resource languages, but with limited capacity, and can also decrease performance in higher-resource languages due to capacity dilution. The commonality of each language plays an important role in cross-lingual transfer learning.

**Acoustic Indonesian ASR**. There are several studies related to Indonesian ASR on traditional models or acoustic models. Among them is the research conducted by Lestari, Iwano, & Furui [14] where they conducted research related to the creation of an ASR system which they called LVCSR (Large Vocabulary Continuous Speech Recognition) using the basis of Hidden Markov Models (HMMs) and n-gram language models. They used a dataset that they collected themselves by recording native speakers (11 males and 9 females) in which each speaker would read out 328 sentences of text. The dataset they collected has developed into a dataset that will be referred to as TITML-IDN. They used the HTK (Hidden Markov Model Toolkit) version 3.2 [15] as an acoustic model training tool and the CMU-Cambridge SLM (Carnegie Mellon Statistical Language Modeling) Toolkit version 2.0 [16] was also used to train 2-gram and 3-gram language models. They conducted language model training using the Good-Turing back off smoothing technique. The result of their research is the creation of an acoustic model of ASR, namely LVCSR (Large Vocabulary Continuous Speech Recognition) with an average word accuracy of 78.2%. However, by adapting speakers using the MAP technique, the accuracy increases by an average of up to 82.3%. Many have developed ASR using the LVCSR model, one of which is the research conducted by Sakti, et al. [17], where they developed the LVCSR model within the A-STAR project. The A-STAR project itself is an Asian consortium that is expected to advance the state-of-the-art multilingual human-machine interfaces within the Asian region. Research conducted by Sakti, et al. [17] produced an optimum word accuracy of 92.47% in the speech corpus BTEC (Basic Travel Expression

Corpus) task consisting of 42 speakers (20 males and 22 females), and each speaker uttered 510 BTEC sentences.

**Recent Indonesian ASR**. Developments related to ASR, especially the Indonesian language itself, are not as fast as other branches of science, this is because the dataset related to the Indonesian language itself is still limited. Recent studies include those conducted by Prakoso, Ferdiana, & Hartanto [18], where conducted an experiment using the CMUSphinx Toolkit with a limited dataset in their Indonesian ASR development. The basis of the model they use is the HMM using a digit corpus and their own made language model to train with a limited dataset. The WER results they get in training are 14% and in testing, they get 20% in a 27,764 dB environment. The development of Indonesian ASR does not stop there, the latest is a study conducted by Syahputra & Zahra [19] where they used the end-to-end ASR model, namely wav2vec 2.0 in making Indonesian ASR. They pre-trained and fine-tuned the model and achieved a WER of 21% using Common Voice as a benchmark. Even so, their model is claimed to be able to work well in the same domain and environment with a WER value of 5%. If seen, it can be concluded that the WER results tend to be better in similar domains, but when tested with different domains, the model only produces a poor WER. The performance results of this model will be used as the main comparison in Indonesian ASR.

## 3. APPROACH AND MODEL ARCHITECTURE

XLSR-53 learns cross-lingual speech representations by extending wav2vec 2.0 to the cross-lingual setting. This model's approach learns a single set of quantized latent speech representations that are shared across languages. There are four important elements in XLSR-53: (1) Feature Encoder; (2) Quantization Module; (3) Context Networks; and (4) Pre-training & Contrastive Loss. Each element will be explained in detail.

### 3.1 Feature Encoder

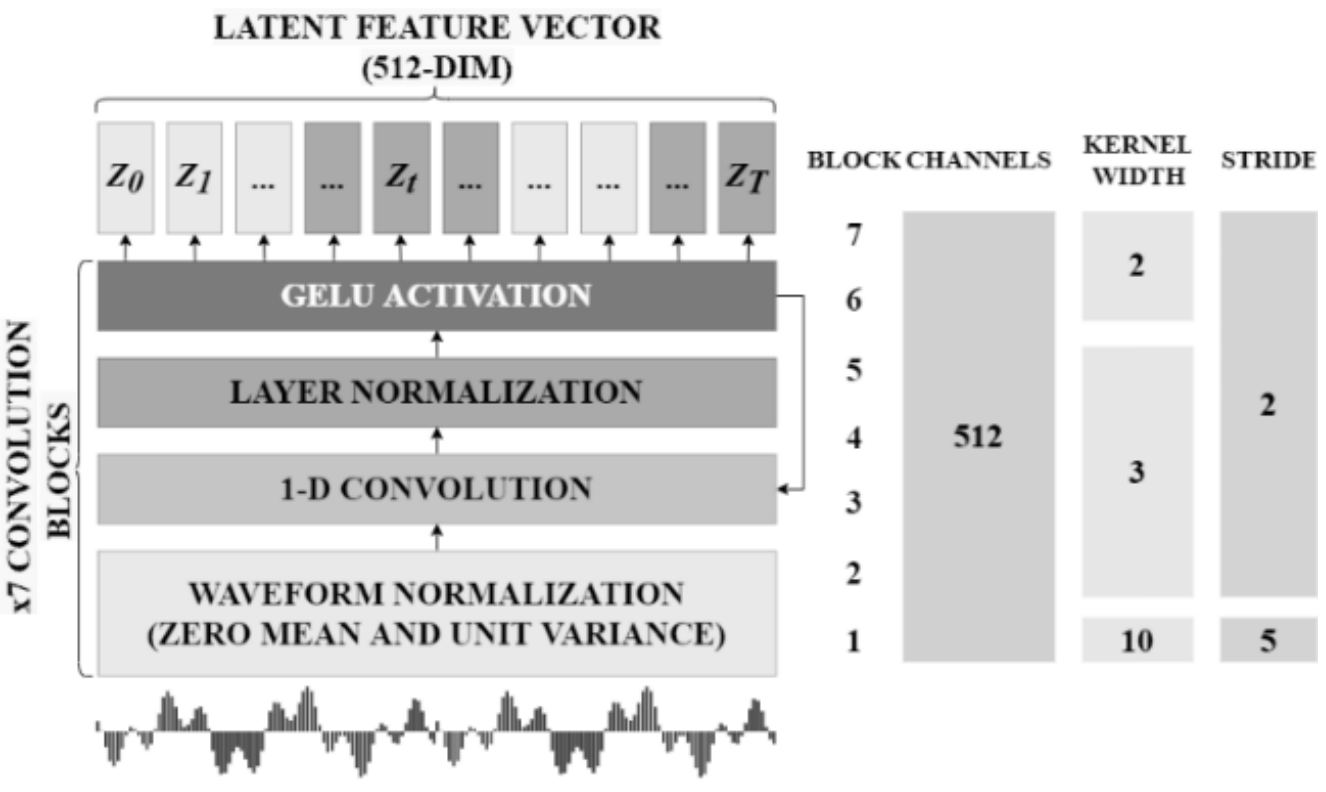

**Figure 4.** Latent Feature Encoder

The task of the feature encoder is to reduce the dimensions of the audio data, the raw waveform is converted into a feature vector sequence Z0, Z1, … , ZT every 20 ms as shown in **Figure 4**. The architecture is simple, which consists of a 7-layer convolutional neural network (single dimension) with 512 channels in each layer with strides [5, 2, 2, 2, 2, 2, 2] and kernel widths [10, 3, 3, 3, 3, 2, 2].

First, the waveform will be normalized before being sent to the network, and the kernel width and stride of the convolutional layers will decrease as the network gets higher. The feature encoder has a total receptive field of 400 samples or 25 ms of audio (audio data is encoded at a sample rate of 16 $KHz$).

### 3.2 Quantization Module

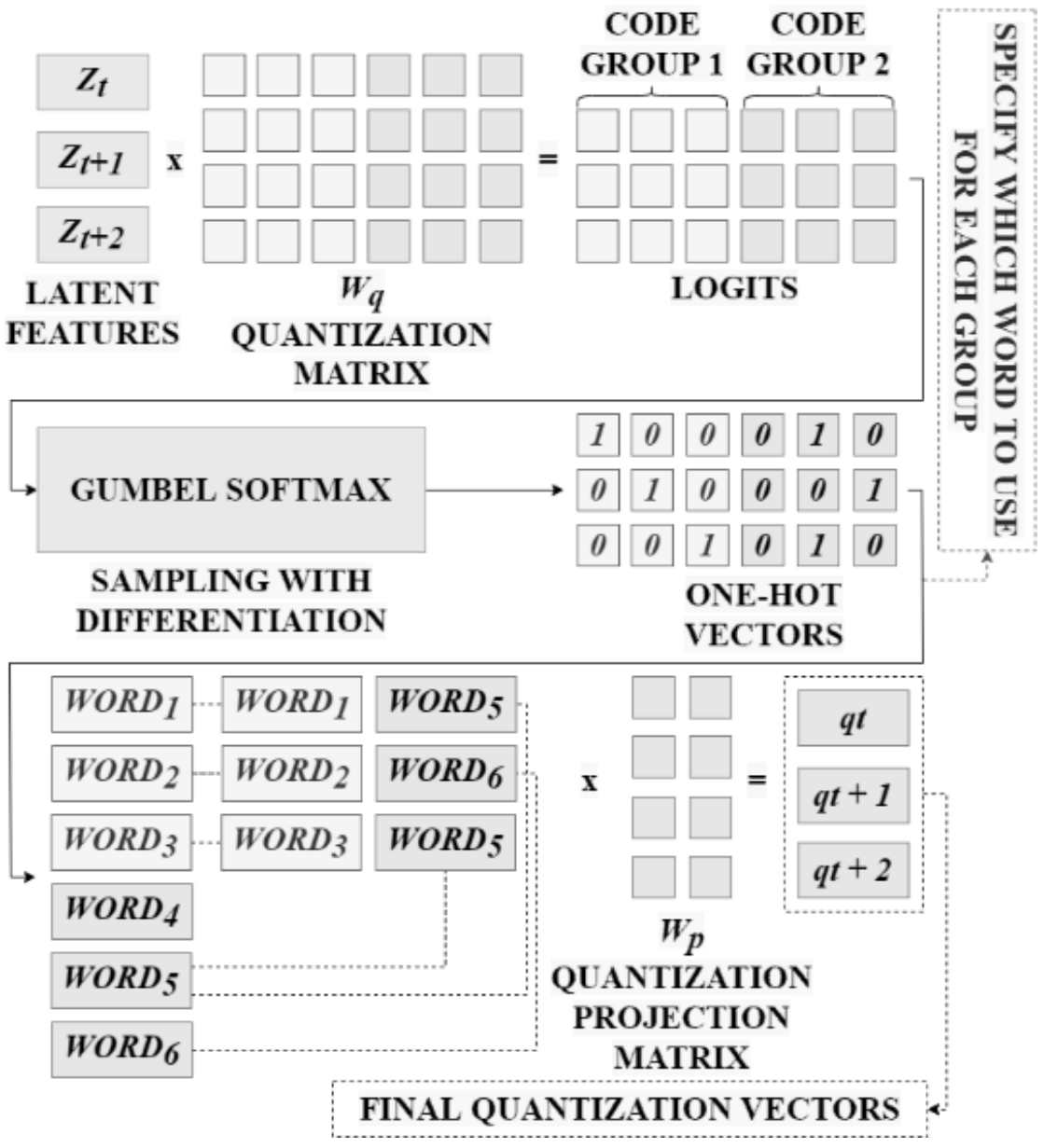

**Figure 5.** Quantization Module

In **Figure 5**, the pre-trained XLSR-53 model can automatically learn discrete speech units by sampling from the Gumbel-Softmax distribution. Units are created from codewords sampled from codebooks (groups). The codewords are then combined to form the final speech unit. This model uses 2 groups with 320 possible words in each group, so theoretically maximum of 320 x 320 = 102,400 speech units.

Latent features are multiplied by the quantization matrix to give logits: one score for each possible codeword in each codebook. The Gumbel-Softmax trick allows sampling one codeword from each codebook, after converting these logits into probabilities. This is similar to taking argmax, the difference being that the operation is fully differentiated. In addition, a small randomness effect, whose effect is controlled by the temperature argument, is introduced to the sampling process to facilitate training and utilization of codewords.

### 3.3 Context Network

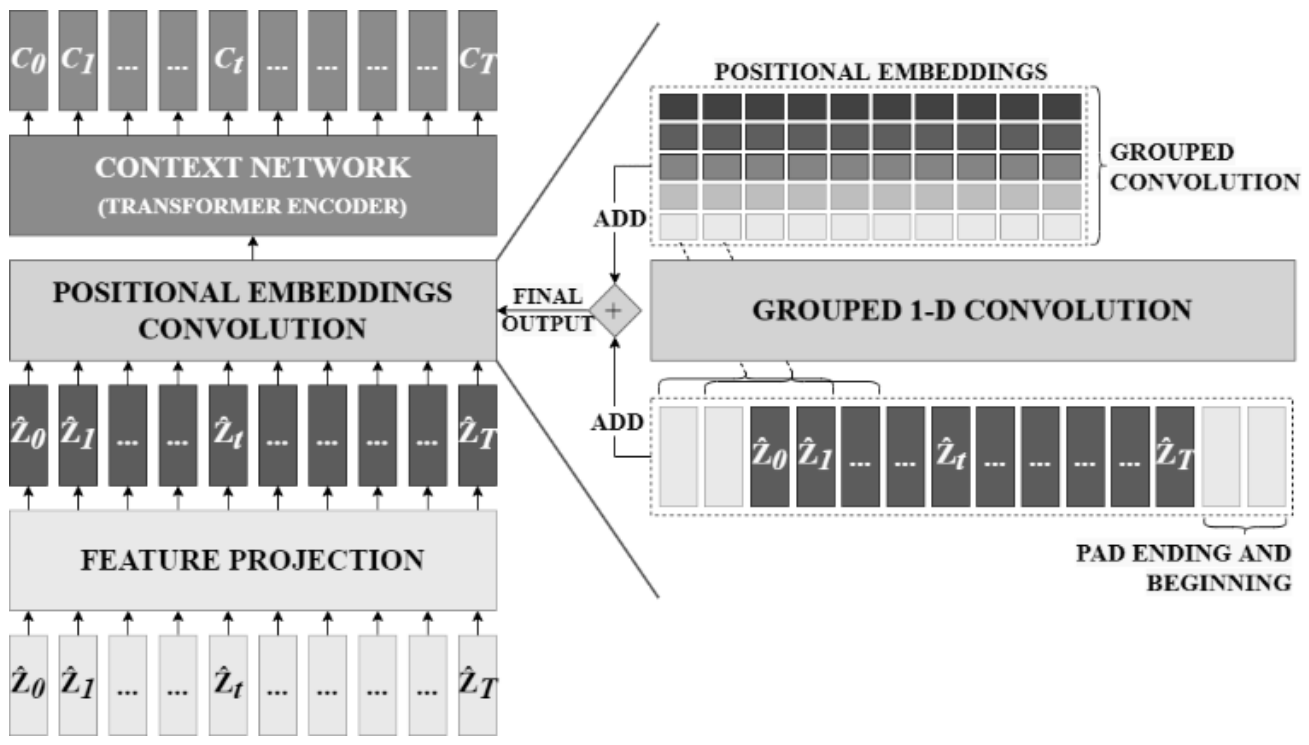

**Figure 6.** Context Network (Transformer Encoder)

One difference from the basic Transformer architecture is how positional information is added to the input. Because Transformer's self-attention operations do not preserve the order of the input sequence, the pre-built fixed embeddings are added to the input vectors in the original implementation. This model uses a new grouped convolution layer to study relative positional embeddings by itself.

The core of this model is its Transformer encoder which can be seen in **Figure 6**, which takes latent feature vectors as input and processes them through 12 Transformer blocks for the BASE version of the model or 24 blocks for the LARGE version. To match the inner dimensions of the Transformer encoder, the input sequence must first pass through the feature projection layer to increase the dimensions from 512 (CNN output) to 768 for BASE or 1,024 for LARGE.

### 3.4 Pre-training & Contrastive Loss

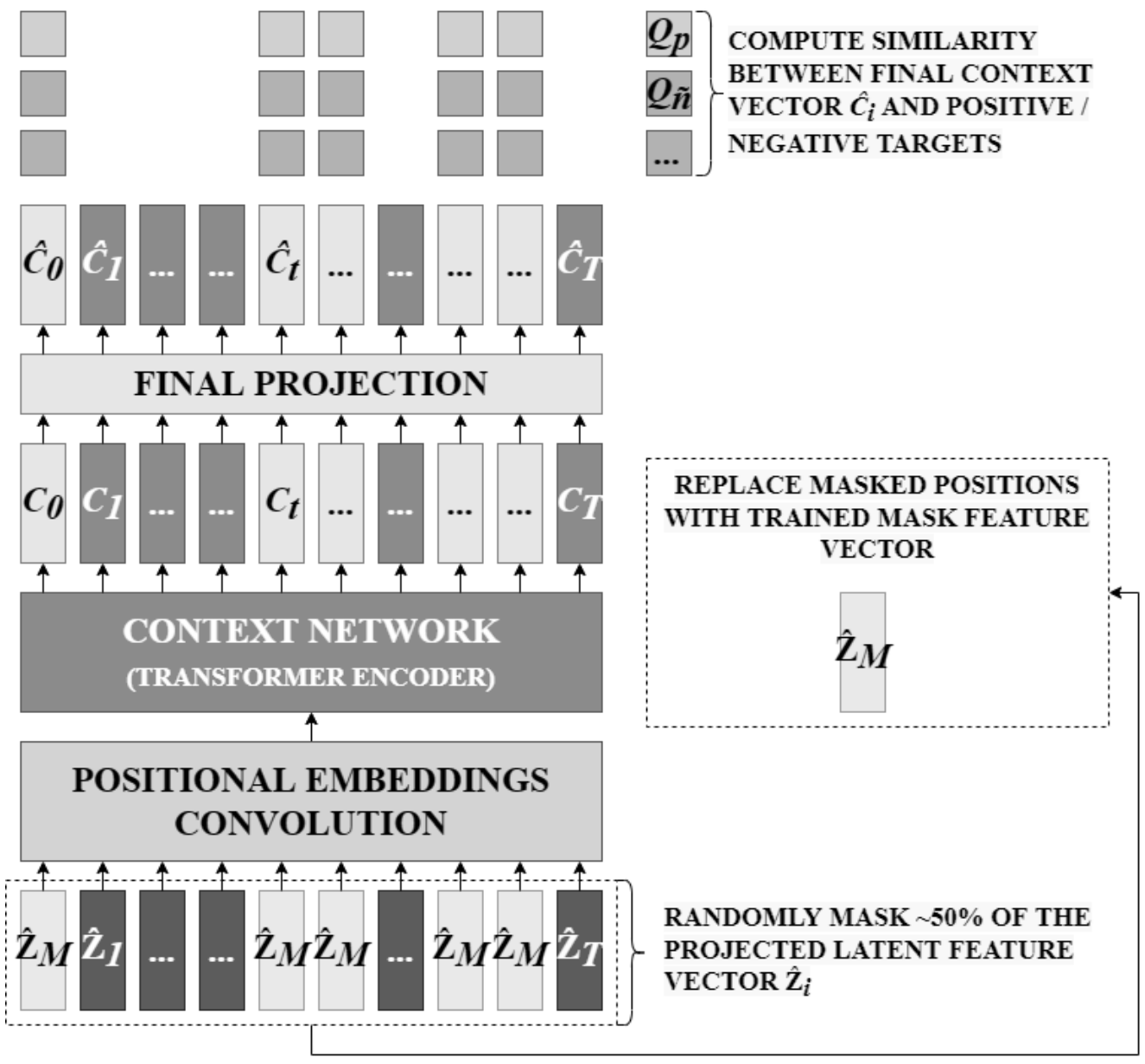

**Figure 7.** Contrastive Loss

The pre-training process uses contrastive tasks to train unlabeled speech data. In **Figure 7**, masks are applied randomly in latent space, where ~50% of the projected latent feature vectors. Masked positions are then replaced by the same trained vector $\hat{Z}_M$ before being fed into the Transformer network.

The final context vectors will then pass through the last projection layer to match the dimensions of the quantized speech units $Q_t$. The quantization is based on product quantization, with quantized representations chosen from $G$ = 2 codebooks, each with $V$ = 320 entries. For each masked position, 100 negative distractors were sampled uniformly from other positions in the same sentence. The model then compares the cosine similarity between the projected context vector t and the true positive target $Q_p$ along with all negative distractors $Q_{\tilde{n}}$. Contrastive loss then encourages high similarity with true positive targets and penalizes high similarity scores with negative distractors.

During pre-training, another loss is added to the contrastive loss to encourage the model to use all codewords equally. It works by maximizing the entropy of the Gumbel-Softmax distribution, preventing the model from always selecting from a small sub-group of all available codebook entries.

## 4. EXPERIMENT

### 4.1 Dataset

Three datasets are used in this study, TITML-IDN, Magic Data and Common Voice. For evaluation and benchmarking in performance comparisons with the Indonesian ASR model that was carried out by Syahputra & Zahra [19], a test split in the Common Voice dataset is used. The total amount of data used in this study is 24 hours, 18 minutes, and 1 second: (1) TITML-IDN contains 14 hours and 31 minutes; (2) Magic Data contains 3 hours and 33 minutes; and (3) Common Voice contains 6 hours, 14 minutes, and 1 second.

**TITML-IDN**. This dataset contains Indonesian speech data from 20 speakers consisting of 11 males and 9 females. Each speaker was asked to read 343 phonetically balanced sentences selected from the text corpus. Most of the sentences were extracted from the text corpus for text-based Information Retrieval (IR), and some sentences were added to the list. The text corpus developed by the Information and Language Processing System (ILPS) group is open to the public. The articles in the corpus are taken from the websites of two popular Indonesian newspapers and magazines. Then the sentences with the following criteria were chosen:

1. Sentences with a relatively large number of different recognition units (phone and biphone) in the text corpus are selected first. At that time, 1,000 sentences were selected.
2. From the 1,000 sentences, sentences consisting of units with a lower frequency of occurrence are selected compared to other units in other sentences which have higher priority. Then extracted about 300 sentences.
3. Added another sentence (43 sentences) which he made by himself including biphone which is not in the corpus and including the unit with a higher frequency of occurrence in the corpus.

Additional information:

1. Number of sentences per speaker : around 343
2. Number of words generated : 4,687
3. Number of phones : 23,570
4. Number of diphones : 18,821
5. Average number of phones per word : 5.0
6. Average sentence length : 14.3 words

The 20 speakers are from the 5 largest ethnic groups in Indonesia: (1) Javanese; (2) Sundanese; (3) Madura; (4) Minang; and (5) Batak. The sound is recorded in a quiet room (clean) using a DAT (Digital Audio Tape) cassette. The equipment used for recording was a Sony DAT-recorder DTC-2000ES and a Sennheiser HMD 25-1 microphone. The recordings are transferred to a file at a sampling rate of 16 $KHz$ with .wav audio format. The normal recording time for one speaker is about 5 minutes, and the total time of the speech corpus after sentence segmentation is 14.5 hours. This dataset was created by Lestari, Iwano, & Furui [14] in their research related to the creation of an ASR system which they called LVCSR.

**Magic Data**. Beijing Magic Data Technology Company Limited (Magic Data Tech) is a global AI (Artificial Intelligence) data services provider headquartered in Beijing. Magic Data Tech provides professional data services for companies and academic institutions involved in artificial intelligence R&D (Research and Development) and application research for speech recognition/ASR, speech synthesis (TTS), and Natural Language Processing (NLP).

**Table 1.** The details about the Indonesian Scripted Speech Corpus – Daily Use Sentence dataset

| SPEAKER_ID | GENDER | AGE | REGION | DEVICE |
|---|---|---|---|---|
| G0004 | M | 26 | Jakarta | HONORRVL-AL09 |
| G0005 | M | 25 | Jakarta | XiaomiRedmi Note 4 |
| G0006 | M | 28 | Jakarta | xiaomiMi A1 |
| G0007 | M | 29 | Jakarta | xiaomiRedmi Note 5 |
| G0008 | M | 21 | East Java | HUAWEIBLA-AL00 |
| G0111 | F | 20 | Sulawesi | samsungSM-J730G |
| G0112 | F | 25 | Jakarta | OPPOA37f |
| G0113 | F | 21 | Jakarta | OPPOA37f |
| G0114 | F | 23 | Bali | motorolaXT1663 |
| G0115 | F | 23 | Jakarta | samsungSM-J710F |

To be exact, the dataset in this study is using "Indonesian Scripted Speech Corpus – Daily Use Sentence" from MagicHub. MagicHub itself is an open-source data platform community developed by Magic Data Tech dedicated to assisting AI developers in model training and promoting the development of an open-source ecosystem. This dataset is not clean and consists of 3.5 hours of transcribed Indonesian scripted speech focusing on daily use sentences, where 10 speakers pronounce 3,296 utterances. The following are the details of the dataset which can be seen in **Table 1**. Magic Data has a sampling rate of 16 $KHz$, so it doesn't need to resample the data.

**Common Voice**. Common Voice itself is a multilingual corpus consisting of more than two million hours of speech data in 38 languages [20]. It is a crowdsourcing project started by Mozilla to create a free useful database for speech recognition software from various languages. This project is supported by volunteers who record various sentences using a microphone and review recordings from other users. The transcribed sentences will be collected in a voice database which is available under a CC0 public domain license. This license aims to ensure that developers can use the database for voice-to-text applications without restriction or cost.

Common Voice used in this study was obtained from Hugging Face whereas Hugging Face used Common Voice version 6.1. The dataset in Common Voice contains voices that are not clean, this is because the sound is taken from various devices, environments, and speakers. The sound recording on Common Voice has .mp3 format so it is necessary to convert the data into .wav. The sampling rate for Common Voice is 48 $KHz$, so it takes a resample of the data to be 16 $KHz$.

This dataset by default has been divided into 6 subsets: (1) train; (2) validation/dev; (3) tests; (4) other; (5) validated; and (6) invalidated and only 3 subsets are used in this study, namely train, validation/dev, and test. The number of speech data in the three subsets in this dataset is 5,809 pieces of data with a total of 6 hours, 14 minutes, and 1 second. In it, there are 170 speakers with varying durations.

The details of the distribution of subsets in this dataset have been divided automatically with the distribution of the train of 36.37%, the validation of 31.59%, and the test of 31.74%. The percentage of this division is based on the amount of data in each subset where the train has 2,130 pieces of data, the validation has 1,835 data, and the test has 1,844 data. However, in this study, the train and validation subsets will be combined first so that these two subsets become one unit (68.26% or 3,965 pieces of data), and then this subset unit is divided again by 90% division (as many as 3,569 pieces of data) for train and 10% (396 pieces of data) for validation.

Data contains:
- Male by 62%
- Female by 38%

With age:
- < 19 : 39%
- 19 – 29 : 43%
- 30 – 39 : 11%
- 40 – 49 : 5%
- 50 – 59 : 2%

### 4.2 Experimental Stages

The stages are divided into four stages: (1) Initiation; (2) Execution; (3) Build Language Model; and (4) Evaluation. Those four experimental stages can be seen in the big picture in **Figure 8**, the details will be explained.

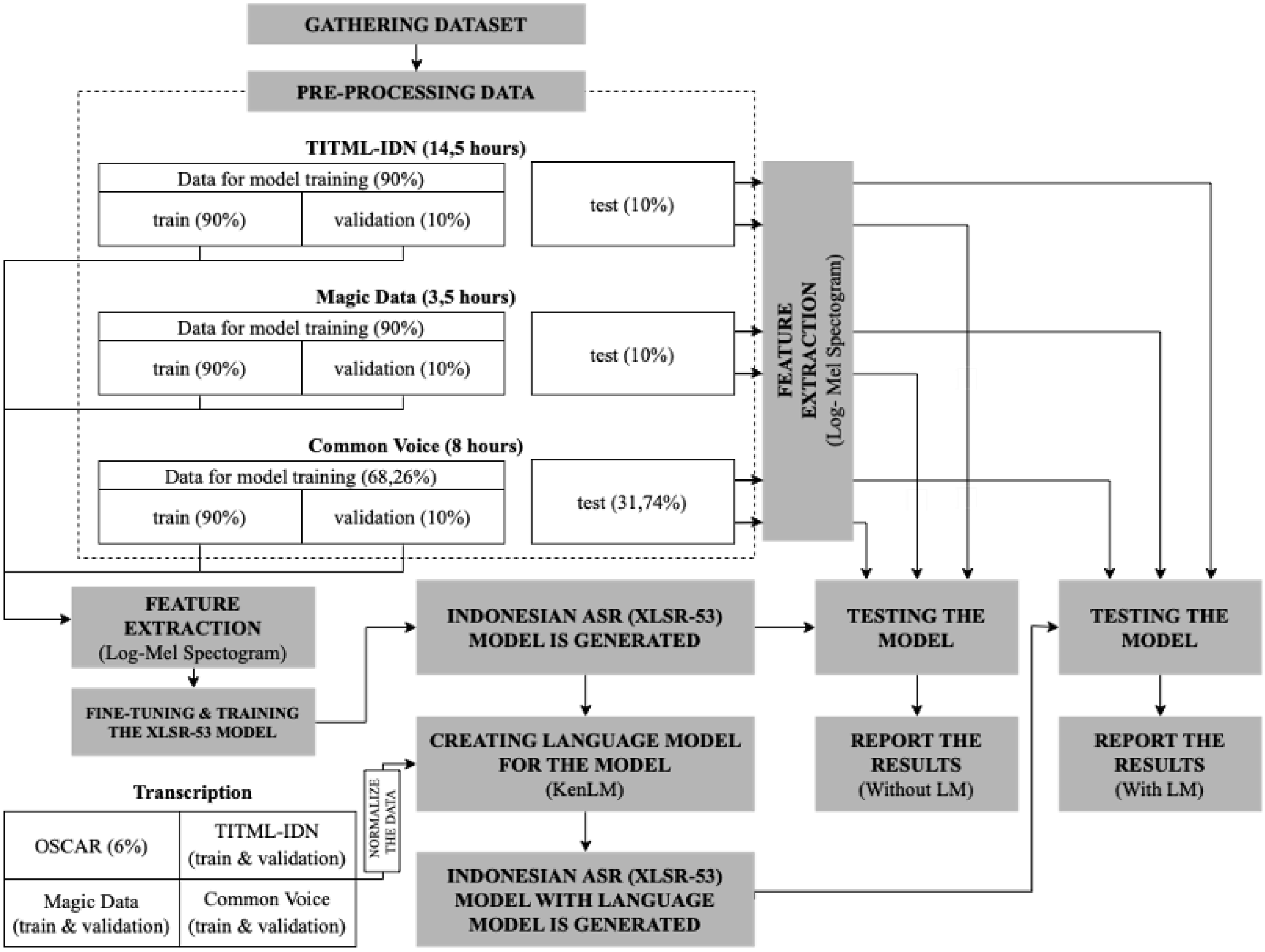

**Figure 8.** Proposed Methodology

**Initiation**. When the data has been obtained, pre-processing is carried out on the data, where the three datasets were combined, tidied up, and adjusted to transcription. For the Common Voice dataset, it is necessary to convert the format from .mp3 to .wav with a single channel specification and a sampling rate of 16 $KHz$. Then normalization is performed, special characters in transcription will be removed in the datasets so that only the alphabet remains and contains only lowercase letters. At this stage, all the different letters are extracted from the train and validation data and build vocabulary from this collection of letters. The mapping function is written so that it is possible to combine all transcriptions into one long transcription and then convert the string into a set of chars. It's important to pass the batched = True argument to the map(…) function later so that the mapping function has access to all transcriptions at once. Then combine all the different letters in the train and validation data and convert the resulting list into an enumerated dictionary. Finally, to make it clearer that “ ” has its token class, it will be changed to a character that looks more like “|”, in addition, an “unknown” token will be added so that the model can later handle characters not found in the training set. Token padding will be added that is adjusted to the "blank token" in the CTC (Connectionist Temporal Classification) algorithm. The "blank token" is a core component of the CTC algorithm. When the data has obtained the number of tokens, the tokenizer can be created by calling the Wav2Vec2CTCTokenizer class in the Hugging Face Transformers library.

**Execution**. A processor will be created which contains a tokenizer that works to process the output format of the model into text form and a feature extractor that works in processing the raw sound signal into the input format of the model, for example, feature vector. At the Initiation stage, the data has been prepared until the tokenizer is created, then feature extraction will then be carried out, where in this XLSR-53 pre-trained model there is a feature extractor module that allows the model to process 16 $KHz$ raw voice signals. It was also found that subsequent layers of the feature encoder can encode all features into more dense and informative representation vectors without intelligent processing of the model. This stage must be done before entering the fine-tuning stage using the CTC algorithm.

Before making a feature extractor, the data must be discretized, which is usually called sampling. Thus, the sampling rate plays an important role in determining the number of data points of the voice signal measured per second. The XLSR-53 pre-trained model expects its input data to have been sampled more or less from the same distribution as the pre-training data. If the same voice signal has been sampled at two different rates, it will produce very different distributions. The pre-trained model of XLSR-53 was pre-trained using Common Voice, BABEL, and Multilingual LibriSpeech datasets, most of which were sampled at 16 $KHz$. So, no matter what train data is used later, make sure it's done by downsampling or upsampling to 16 $KHz$.

The feature extractor in the XLSR-53 pre-trained model can be created by calling the Wav2Vec2FeatureExtractor class in the Hugging Face Transformers library and requires the following parameters in **Table 2** for operation.

**Table 2.** Parameters that must be defined when calling the Wav2Vec2FeatureExtractor class in the Hugging Face Transformers library

| | |
|---|---|
| *feature_size* | Feature dimension of extracted feature |
| *sampling_rate* | The sampling rate at which audio files must be digitized is expressed in Hertz per second (Hz) |
| *padding_value* | The value used to fill the padding value |
| *do_normalize* | Whether the zero-mean unit-variance normalizes the input or not. Normalization can help significantly improve performance for some models |
| *return_attention_mask* | Should the model use attention_mask for batched inference? In general, the XLSR-53 pre-trained model should always use the attention_mask attribute |

With the feature extractor made, the criteria needed for making the processor are finished. Furthermore, the tokenizer and feature extractor will be combined into one class Wav2Vec2Processor, which is the processor for the model.

Next, fine-tuning will be carried out in two scenarios: (1) with a language model; and (2) without a language model. These two scenarios are carried out to provide an evaluation of how significant the performance is when the language model is used on the XLSR-53 pre-trained model. When fine-tuning, this model has added a classifier that represents the output vocabulary of each downstream task on top of the model and performs training on labeled data using CTC loss [21, 22]. During fine-tuning, the weight of the encoder feature is not updated. Based on the dev set error rate in the study of Conneau, et al. [7], the best learning rate is in [2e-5, 6e-5]. Instead, this study using a learning rate in [3e-4] with batch_size = 12 and epoch = 20. Evaluation is carried out every 200 steps and a save checkpoint will be carried out every 200 steps as well, save checkpoints are limited to a maximum of the last 2 checkpoints so that when after the third save checkpoint, the first checkpoint will be deleted and so on. In addition, gradient_accumulation_steps in this study = 2.

The first scenario will be decoded using the 5-gram KenLM language model. To create a 5-gram language model, KenLM [23] was used with a text corpus from four sources (OSCAR, TITML-IDN, Magic Data, and Common Voice). The fine-tuning process uses Google Colab Pro.

**Build Language Model**. After the model has been built and before entering the evaluation stage, a language model will be created first. Making a language model requires a text corpus that is larger than the text transcript in the speech dataset so that the language model can have the right probability value for each word in the corpus. The text used in the process of making the language model only consists of alphabetic characters, spaces, and only lowercase letters obtained through the normalization process. As previously mentioned that to create a 5-gram language model, KenLM was used with a text corpus from four sources: (1) OSCAR (only used 6%); (2) TITML-IDN; (3) Magic Data; and (4) Common Voice.

OSCAR (Open Super-large Crawled ALManaCH coRpus) is an additional text corpus in making this language model, this is necessary because OSCAR is a text corpus that is quite large so that it can affect the results of the language model. There are 2,394,957,629 Indonesian words in the "unshuffled_deduplicated_id" subset, with such large data, only about 6% of the transcription data was used.

The results of making the language model are saved in the form of an .arpa file. The resulting language model format consists of \data\ which gives multiple entries for each N-gram. The next N section gives the words and pairs found in the text corpus used. Each N-gram row consists of probability values in the form of log10, words or pairs, and backoff weight values. Lastly, the language model obtained using KenLM is converted into a binary file to reduce the size of the N-gram and make loading faster, KenLM allows converting .arpa files to binary ones using the build_binary executable.

**Evaluation**. After the two scenarios in the Execution stage have given results, an evaluation is carried out using the TITML-IDN, Magic Data, and Common Voice datasets. Testing using the Common Voice dataset is intended to be a benchmark for the modeling carried out by Syahputra & Zahra [19], where they also use the test split in the Common Voice dataset as testing data. In contrast to the study of Syahputra & Zahra [19] whose training data uses an internal dataset of a company, namely BahasaKita, this study uses TITML-IDN, Magic Data, and Common Voice with train and validation split as training data. Measurement of the model using the WER metric.

## 5. RESULT

As seen in

**Table 3**, the model produced in this study can compete with similar models carried out by Syahputra & Zahra [19] when using a test split of the Common Voice dataset. The model produced by Syahputra & Zahra [19] obtained a WER value of 21% when the language model was not included, whereas our study has managed to provide a better WER of 20%. In addition, when the language model is included, the model in Syahputra & Zahra [19] research obtains a WER value of 41% while ours is 12%.

It should be noted that the Indonesian ASR model by Syahputra & Zahra [19] went through a pre-training process with a data duration of 345 hours before finally entering the fine-tuning process, this step is necessary because wav2vec 2.0 is an English-based model, while in this study did not go through the pre-training process and went straight into the fine-tuning process with train and validation split data, it is because the XLSR-53 pre-trained model has been pre-trained with 53 different languages, including Indonesian. Syahputra & Zahra [19] also divided their research into two scenarios when fine-tuning the model: (1) Low resource fine-tuning with a dataset of 25 hours; and (2) High resource fine-tuning with a dataset of 75 hours.

The use of the language model can improve the WER with a very significant improvement value. The main key when creating a language model is the transcription used, if the transcription is large enough, the resulting language model will have better quality. In addition, the length of the context in the language model can also affect the quality of the resulting language model. The longer the context (the greater the value of n in n-grams), then when it is embedded into the ASR model, the better the performance generated by the model. But when there are too many contexts, it doesn't mean the performance will be better.

**Table 3.** Comparison of evaluation results with similar models. The percentage are using WER metric

| Model | Data Training & Validation | Language Model | TITML-IDN | Magic Data | Common Voice | BahasaKita – batch 10 | AVG WER |
|---|---|---|---|---|---|---|---|
| **THIS STUDY** | | | | | | | |
| Indonesian ASR (XLSR-53) | TITML-IDN (14j 31m) | - | 1.611% | 39.291% | 48.946% | – | 29.950% |
| | | 2-gram KenLM | 0.662% | 28.709% | 35.903% | – | 21.758% |
| | | 3-gram KenLM | **0.612%** | 28.253% | 35.903% | – | 21.590% |
| | | 4-gram KenLM | **0.612%** | 28.253% | 35.866% | – | 21.577% |
| | | 5-gram KenLM | **0.612%** | 28.253% | 35.875% | – | 21.580% |
| | Magic Data (3j 33m) | - | 38.776% | 20.405% | 47.548% | – | 35.576% |
| | | 2-gram KenLM | 17.239% | 12.304% | 31.661% | – | 20.401% |
| | | 3-gram KenLM | 15.690% | 12.152% | 31.624% | – | 19.822% |
| | | 4-gram KenLM | 15.453% | 12.304% | 31.624% | – | 19.794% |
| | | 5-gram KenLM | 15.453% | 12.304% | 31.652% | – | 19.803% |
| | Common Voice (6j 14m 1d) | - | 42.961% | 33.873% | 26.123% | – | 34.319% |
| | | 2-gram KenLM | 24.122% | 22.025% | 15.215% | – | 20.454% |
| | | 3-gram KenLM | 22.786% | 21.215% | 15.458% | – | 19.820% |
| | | 4-gram KenLM | 22.486% | 21.519% | 15.290% | – | 19.765% |
| | | 5-gram KenLM | 22.486% | 21.620% | 15.243% | – | 19.783% |
| | TITML-IDN + Magic Data (18j 4m) | - | 2.286% | 16.152% | 38.216% | – | 18.885% |
| | | 2-gram KenLM | 1.012% | 11.443% | 25.247% | – | 12.567% |
| | | 3-gram KenLM | 0.937% | 11.291% | 25.200% | – | 12.476% |
| | | 4-gram KenLM | 0.937% | 11.342% | 25.284% | – | 12.521% |
| | | 5-gram KenLM | 0.937% | 11.342% | 25.284% | – | 12.521% |
| | TITML-IDN + Common Voice (20j 45m 1s) | - | 2.473% | 27.392% | 21.070% | – | 16.979% |
| | | 2-gram KenLM | 1.149% | 20.354% | 13.435% | – | 11.646% |
| | | 3-gram KenLM | 1.012% | 20.152% | 13.388% | – | 11.517% |
| | | 4-gram KenLM | 1.012% | 20.101% | 13.388% | – | 11.500% |
| | | 5-gram KenLM | 1.012% | 20.203% | 13.435% | – | 11.550% |
| | Magic Data + Common Voice (9j 47m) | - | 34.354% | 16.911% | 23.112% | – | 24.792% |
| | | 2-gram KenLM | 14.154% | 11.190% | 13.957% | – | 13.100% |
| | | 3-gram KenLM | 12.242% | 10.785% | 13.938% | – | 12.322% |
| | | 4-gram KenLM | 12.205% | **10.684%** | 13.891% | – | 12.260% |
| | | 5-gram KenLM | 12.155% | **10.684%** | 13.975% | – | 12.271% |
| | TITML-IDN + Magic Data + Common Voice (24j 18 m 1 d) | - | 2.174% | 16.759% | 20.306% | – | 13.080% |
| | | 2-gram KenLM | 0.775% | 10.785% | 12.232% | – | 7.930% |
| | | 3-gram KenLM | 0.725% | 10.886% | 12.297% | – | 7.969% |
| | | 4-gram KenLM | 0.725% | 10.886% | **12.213%** | – | **7.941%** |
| | | 5-gram KenLM | 0.725% | 10.937% | 12.251% | – | 7.971% |
| **BENCHMARK** | | | | | | | |
| Indonesian ASR by Syahputra & Zahra (2021) (wav2vec 2.0) | BahasaKita batch 11 (25j) | – | – | – | 24.000% | 7.000% | 15.500% |
| | | 3-gram KenLM | – | – | 45.000% | 26.000% | 35.500% |
| | BahasaKita batch 10 – 12 (75j) | – | – | – | 21.000% | 5.000% | 13.000% |
| | | 3-gram KenLM | – | – | 41.000% | 22.000% | 31.500% |

In

**Table 3**, the WER from the evaluation results on the TITML-IDN test split has very good results, it can even touch numbers below 1%. After being traced, this is presumably because the sentences used in the TITML-IDN dataset have not so many variations, so the model can predict them easily in the testing data because the testing data is thought to have the same sentences in the training data even though the speakers are different.

## 6. CONCLUSION AND FUTURE WORK

The need for speech recognition technology in Indonesia has to be intensified, but with the limited dataset related to the Indonesian language itself, speech recognition technology in Indonesia is a little behind.

The model used in this study is the pre-trained XLSR-53 model, which this model is proven to exceed the previous ASR model (wav2vec 2.0). The model is fine-tuned using datasets from TITML-IDN, Magic Data, and Common Voice with a total duration of 24 hours, 18 minutes, and 1 second divided by the train, validation, and test subsets.

With this study, Indonesian ASR has been built with good accuracy even though it only uses fewer datasets, it is hoped that it can contribute to speech recognition in Indonesia. The following conclusions can be summarized:

1. It can be concluded that the use of the XLSR-53 pre-trained model in conducting cross-lingual transfer learning which is implemented in Indonesian can give good results. This is evident when comparing the pre-trained XLSR-53 model, this model can produce a better WER performance than the previous model, namely wav2vec 2.0 in learning Indonesian even by using fewer data.

2. The more varied or the more datasets used when retraining and fine-tuning, the better the performance produced by the XLSR-53 pre-trained model. This is evidenced when the model is trained with various types of datasets (TITML-IDN, Magic Data, and Common Voice), and the model can produce a fairly good WER.
3. The use of a language model is very important because the language model uses a corpus from the literature that has been formed to calculate the possibility of a series of words or sentences. The language model is used to solve sentence syntax in speech recognition. When creating a language model, we need a text corpus that is larger than the text transcript in the speech dataset so that the language model can have the correct probability value for each word in the corpus [24].
4. The model produced in this study can compete with the previous study using a subset test from the Common Voice dataset with a WER of 20% (without a language model) and WER can be reduced by approximately 8% using a language model to 12%.

There is still much that can be done regarding this study. The proposed model should be trained with more diverse data to maximize the potential that exists in the model. The data in question may have different dialects, different accents, different speaker identities, different emotional states, different recording devices, or be recorded in different environments so that the model can learn various variations. In addition, with the limited resources provided by Google Colab Pro, it would be better if the model was trained using more qualified resources.

It is also felt that the model will provide better performance if the signal is processed, such as applying the Hilbert-Huang transform to filter out noise in the speech data first. As has been done by Jiang, N., & Li, J. [25] where the simulation results from their research show that the algorithm they propose can remove most of the noise in the voice signal by using the Hilbert-Huang transform.

On the other hand, judging from the ability of the XLSR-53 pre-trained model that can share latent speech representation between languages and the possibility to use language features from other languages without having to re-train the model in a new language, it would be more interesting if this model trained to be able to learn various languages other than Indonesian. Considering that Indonesia itself has a lot of scattered regional languages. That way, the voice recognition technology offered to the Indonesian people will feel more generalized.